\documentclass[journal]{IEEEtran}
\usepackage{graphicx}
\usepackage{amsfonts}

\usepackage[english]{babel}
\usepackage{amsmath, amssymb}
\usepackage{float}
\usepackage{makecell}
\usepackage{multirow}
\usepackage{cite}
\usepackage{amsmath,amssymb,amsfonts}
\usepackage{algorithmic}
\usepackage{graphicx}
\usepackage{textcomp}
\usepackage{xcolor}

\usepackage{subfigure}
\usepackage{epstopdf}

\usepackage{graphicx}
\usepackage{comment}
\usepackage{amsmath,amssymb} 
\usepackage{color}
\usepackage{enumerate}

\usepackage{booktabs}

\ifCLASSINFOpdf
\else
\fi
\begin{document}
%
\title{Hierarchical Attention Learning of Scene Flow \\in 3D Point Clouds}
%
%
%

\author{Guangming~Wang,
       Xinrui Wu, Zhe Liu, and Hesheng Wang
        
\thanks{*This work was supported in part by the Natural Science Foundation of China under Grant U1613218 and 61722309. The first two authors contributed equally. Corresponding Author: Hesheng Wang.}
\thanks{G. Wang, X. Wu, and H. Wang are with Department of Automation, Shanghai Jiao Tong University, Shanghai 200240, China and Key Laboratory of System Control and Information Processing, Ministry of Education of China.}
\thanks{Z. Liu is with the Department of Computer Science and Technology, University of Cambridge.}

}

%
%

\markboth{Journal of \LaTeX\ Class Files,~Vol.~14, No.~8, August~2015}%
{Shell \MakeLowercase{\textit{et al.}}: Bare Demo of IEEEtran.cls for IEEE Journals}
%



\maketitle

\begin{abstract}
Scene flow represents the 3D motion of every point in the dynamic environments. Like the optical flow that represents the motion of pixels in 2D images, 3D motion representation of scene flow benefits many applications, such as autonomous driving and service robot. This paper studies the problem of scene flow estimation from two consecutive 3D point clouds. In this paper, a novel hierarchical neural network with double attention is proposed for learning the correlation of point features in adjacent frames and refining scene flow from coarse to fine layer by layer. The proposed network has a new more-for-less hierarchical architecture. The more-for-less means that the number of input points is greater than the number of output points for scene flow estimation, which brings more input information and balances the precision and resource consumption. In this hierarchical architecture, scene flow of different levels is generated and supervised respectively. A novel attentive embedding module is introduced to aggregate the features of adjacent points using a double attention method in a patch-to-patch manner. The proper layers for flow embedding and flow supervision are carefully considered in our network designment. Experiments show that the proposed network outperforms the state-of-the-art performance of 3D scene flow estimation on the FlyingThings3D and KITTI Scene Flow 2015 datasets. We also apply the proposed network to realistic LiDAR odometry task, which is an key problem in autonomous driving. The experiment results demonstrate that our proposed network can outperform the ICP-based method and shows the good practical application ability.
\end{abstract}


%
\IEEEpeerreviewmaketitle

\section{Introduction}
%
%
%
%
\IEEEPARstart{S}{cene} flow represents 3D motion field \cite{vedula1999three} between two consecutive point clouds in a scene that moves completely or partially. It specifically characterizes the motion distance and direction of each 3D point. It is a fundamental low-level understanding of the 3D dynamic scene like the optical flow for 2D image. The scene flow estimation can be widely applied to many fields such as autonomous driving, motion segmentation, action recognition, etc. However, most previous studies \cite{huguet2007variational,pons2007multi,valgaerts2010joint,menze2015object,vogel2013piecewise,vogel20153d,mayer2016large,ma2019deep} focus on obtaining disparity and optical flow from 2D images to represent the scene flow, and use the accuracy of the disparity and optical flow to evaluate the performance of scene flow estimation. However, this is not suitable for raw 3D points input, like the data from LiDAR used in autonomous driving.  

There are some previous studies \cite{dewan2016rigid,ushani2017learning} that are not based on deep learning for 3D point input but demand for the local rigid assumption of the scene. With advances of deep learning on 3D point clouds \cite{qi2017pointnet,qi2017pointnet++,rahman2019recent}, some recent studies \cite{liu2019flownet3d,gu2019hplflownet,wu2019pointpwc} can generate 3D scene flow from two consecutive point clouds without the knowledge of the scene structure or any prior information of the motion. These methods directly input the coordinates of point clouds into neural network to learn scene flow in an end-to-end fashion. FlowNet3D \cite{liu2019flownet3d} builds several downsampling and upsampling layers like PointNet++ \cite{qi2017pointnet++} and uses a new flow embedding layer to extract the correlation between point clouds. The flow embedding feature is then propagated through set conv and set upconv layers to obtain the scene flow. However, the flow embedding layer merely extract the correlation between a specific point in the first frame and several queried points in the second frame, which is called point-to-patch manner in this paper. The point-to-patch manner does not consider the points near the specific point in the first frame in embedding process. HPLFlowNet \cite{gu2019hplflownet} proposes the patch correlation on permutohedral lattice representation and uses proposed CorrBCLs in multiple layers to learn correlations of two consecutive frames. However, the information loss of interpolation on permutohedral lattice limits the performance. 

The patch-to-patch manner proposed in \cite{wu2019pointpwc} also utilizes more information in embedding process. The patch-to-patch manner uses one patch in the first frame to embed features rather than only one point as the point-to-patch manner. And the idea of hierarchical optical flow learning with a coarse-to-fine fashion from PWC-Net \cite{sun2018pwc,sun2019models} is also used in PointPWC-Net \cite{wu2019pointpwc}. However, though patch-to-patch manner expands the received fields, one point has only one direction flowing to the second frame. That means not all information has the same importance. Actually, weighting point features can make more attention lie in the task-related regions and lose less information compared with the max-pooling operation. And the idea of attention mechanism has shown great significance in other visual tasks \cite{zhang2019multi,wang2020region} and 3D point cloud tasks\cite{zhang2019pcan,hu2020randla}. 

PointPWC-Net \cite{wu2019pointpwc} uses the Multi-Layer Perceptron (MLP) to learn the weight from the relative coordinates of points for flow embedding. However, the correlation weight is not decided only by the Euclidean space. For example, one point in the trunk of a tree may search for the points in a car or wall nearby and the points in the top of the tree in the process of flow embedding. However, the points in the nearby car are less related, even though they are nearer to the trunk compared with the points in the top of the tree. So it is also very important to consider feature space in scene flow estimation. Our paper focuses on this and proposes a hierarchical attention learning network for scene flow estimation with two different attentions in each flow embedding. In addition, taking more points as input to estimate the scene flow of fewer points will give more precious structure information of the scene compared with the fewer input points for estimation. Meanwhile, more input points for fewer estimations reduces the consumption of GPU memory in the process of flow refinements compared with the more input points for more point estimations. What’s more, existing works exploit less on the practical applications of the scene flow in 3D point clouds other than FlowNet3D \cite{liu2019flownet3d}, which tried applying the scene flow to the point cloud registration task comparing with Iterative Closest Point (ICP) \cite{besl1992method} and motion segmentation task. We exploit a more difficult task that is LiDAR odometry task in this paper. Our main contributions are as follows:

\begin{itemize}
\item A novel hierarchical attentive network is proposed for scene flow estimation from two consecutive 3D point clouds. In the network, a novel double attentive flow embedding method for two consecutive point clouds is proposed and applied to the hierarchical attention learning of scene flow.
\item The proposed hierarchical attentive network has a novel more-for-less architecture for the balance of precision and resource consumption. And more 3D Euclidean information is used for flow embedding to improve the performance of the network. 
\item The proper layers for flow embedding and flow supervision are carefully considered and demonstrated by ablation studies in our network design. The experiments on synthetic data from FlyingThings3D \cite{mayer2016large} and real LiDAR scans from KITTI \cite{menze2015joint,menze2018object} demonstrated that our method outperforms the-state-of-the-art methods in 3D scene flow estimation by a large margin.
\item Finally, the practical application ability of our scene flow network is demonstrated on the LiDAR odometry task. Our network can outperform the ICP-based method without any adjustment of network structure. 
\end{itemize}

Our paper is orginazed as follows: The Section II is on the related work. Section III prsents the problem definition of scene flow estimation in 3D point clouds. The details about our hierarchical attention learning network are in Section IV. The traning details, evaluation metrics, the experiments compared with state-of-the-arts, ablation studies, and the application on LiDAR odometry are in Section V.

\begin{figure*}[t]
		\vspace{-1mm}
	\centering
	\includegraphics[scale=0.65]{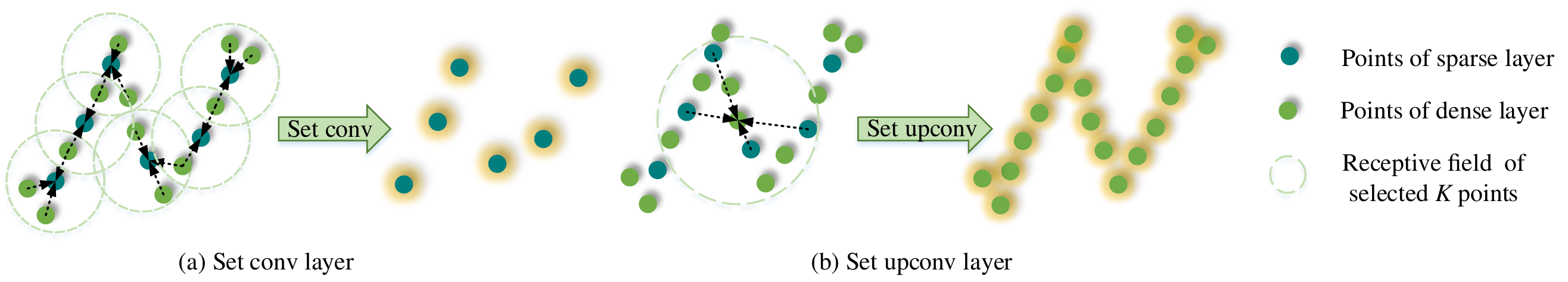}
	\vspace{-2mm}
	\caption{\textbf{The set conv layer and the set upconv layer.} The set conv layer is used in the point feature encoding in Section IV-A and flow feature encoding in Section IV-B. The set upconv layer is used in the flow refinement in Section IV-C.1. In this paper, the sparse points are sampled from the dense points, so the sparse points are part of the dense points.}
	\label{fig:setconv}
\end{figure*}

\section{Related Work}

\subsection{Deep Learning for 3D Data}
As the CNNs has shown great achievements on the image and video input, many previous works convert the raw irregular 3D data to regular data forms, such as voxels \cite{wu20153d,maturana2015voxnet,riegler2017octnet,wang2017cnn,graham20183d,garcia2019geometry} and multi-view \cite{su2015multi,kalogerakis20173d,huang2017learning}. Then the standard Convolutional Neural Networks (CNNs) can be used. However, these methods suffer from the error of data conversion, such as viewpoint selection and the resolution of rendering images or voxels. In order to avoid discretization error, the methods based on point cloud input are high-profile since the pioneer PointNet \cite{qi2017pointnet}, the first work to directly process the point clouds in an end-to-end fashion. PointNet \cite{qi2017pointnet} uses point-wise MLP and max-pooling as the symmetry function to learn the global feature of unordered point clouds for segmentation and classification tasks. Then, PointNet++ \cite{qi2017pointnet++} is proposed for the hierarchical point feature learning by aggregating and learning local feature from local neighborhoods in 3D Euclidean space layer by layer. FlowNet3D \cite{liu2019flownet3d} and our method for scene flow task both use the hierarchical point feature learning of PointNet++ \cite{qi2017pointnet++}.

SPLANet \cite{su2018splatnet} jointly estimates 2D and 3D semantic segmentation in high-dimension permutohedral lattice and adopts bilateral convolutional layer (BCL) \cite{kiefel2014permutohedral,jampani2016learning}. There are also some methods directly handling 3D point clouds based on graphs \cite{wang2019dynamic,jiang2019hierarchical} and 3D meshes \cite{huang2019texturenet,hanocka2019meshcnn,schult2020dualconvmesh}.
 RandLA \cite{hu2020randla} is a recent semantic segmentation method for large-scale 3D point clouds. It utilizes local spatial encoding to explicitly extract the local geometric patterns and uses attentive pooling to replace max-pooling. Our double attentive embedding method is inspired by the attentive pooling \cite{hu2020randla} but with a different task characteristic: 
 the semantic segmentation focuses on classification for each point, while the correlation of two frames of point clouds is the key problem in scene flow task. And we propose a new double attentive embedding method with two different attentions.

\subsection{Scene Flow Estimation}
Many previous works estimate scene flow from RGB stereo \cite{huguet2007variational,pons2007multi,valgaerts2010joint,menze2015object,vogel2013piecewise,vogel20153d,mayer2016large,ma2019deep} and RGB-D data \cite{hadfield2011kinecting,herbst2013rgb,jaimez2015primal}. Some of these works use the variational method with motion smoothness regularizations \cite{huguet2007variational,pons2007multi,valgaerts2010joint}, and some depends on the rigidly assumpings of local pieces \cite{vogel2013piecewise,vogel20153d} or objects \cite{menze2015object,ma2019deep}. These methods are less related with ours based on point cloud input. 

With the advent of commonly used LiDAR in autonomous driving and service robot, the methods estimating scene flow from raw point clouds are developing. Dewan et al. \cite{dewan2016rigid} assume local geometric constancy and formulate the problem of rigid scene flow estimation for LiDAR scans as an energy minimization problem with the factor graph representation. Ushani et al. \cite{ushani2017learning} construct filtered occupancy grids with raw LiDAR data and then estimate locally rigid scene flow through an Expectation-Maximization (EM) algorithm. \cite{behl2019pointflownet} estimates the ego-motion to represent the rigid motion of background and individual rigid motion of objects to composite complete scene flow of scenes. \cite{shao2018motion} learns object segmentation and rigid motions of individual objects. Our method learns soft correspondence from the raw point clouds through an end-to-end fashion and is independent of the rigidity assumptions.

With the development of 3D deep learning on raw point clouds, some methods directly estimate scene flow from raw point clouds. Wang et al. \cite{wang2018deep} propose a learnable parametric continuous convolution and test the convolution on many tasks. They also test on scene flow estimation task but give few details on the implementation, and their test dataset is belonging to Uber and not publicly available. FlowNet3D \cite{liu2019flownet3d} incorporates hierarchical PointNet++ \cite{qi2017pointnet++} and learns the scene flow through a novel flow embedding layer. HPLFlowNet \cite{gu2019hplflownet} proposes DownBCL, UpBCL, and CorrBCL operations to restore and learn the information in permutohedral lattice and estimates the scene flow in large-scale point clouds. But the barycentric interpolation in permutohedral lattice introduces errors \cite{gu2019hplflownet}. We expect to use the flexibility of raw point clouds and do not interpolate points to lattices. PointPWC-Net \cite{wu2019pointpwc} uses patch-to-patch manner and the idea of hierarchical flow refinement from PWC-Net \cite{sun2018pwc,sun2019models} to estimate scene flow. This paper focuses on more efficient point correspondence of two frames of point clouds and proposes a novel double attentive flow embedding method. We also apply the scene flow estimation network to the LiDAR odometry task to show the application ability of our method.

\section{Problem Definition}

In this section, we expound the task of scene flow estimation in 3D point clouds. There are two sets of point clouds in two consecutive frames. The scene flow is a vector indicating how each point in the first frame flows to another location in the second frame, and the task is to estimate the scene flow for each point in the first frame without any assumption. Specifically, our proposed hierarchical attention network receives two sampled sets of point clouds (PC) from two consecutive frames as input. The two sets of point clouds are  $P{C_1} = \{ {x_i} \in {\mathbb{R}^3}|i = 1, \ldots ,{N_1}\} $  at the $t$  frame and $P{C_2} = \{ {y_j} \in {\mathbb{R}^3}|j = 1, \ldots ,{N_2}\} $  at the $t + 1$  frame, where  ${x_i}$ and ${y_j}$  are the 3D coordinates of the points. The network outputs the scene flow $\widehat {SF} = \{ \widehat{sf_i} \in {\mathbb{R}^3}|i = 1, \ldots ,{N_1}\}$  for each point ${p_i}$  in the $P{C_1}$. 

Note that the points in two frames do not need to have the same amount or correspond to each other since the perceived two point clouds are discretely sampled from the environment at different time separately. That is, the predicted $\widehat {PC_2} = \{ {x'_i} \in {\mathbb{R}^3}|{x'_i} = {x_i} + \widehat{sf_i},i = 1, \ldots ,{N_1}\}$ are not expected to correspond to $P{C_2}$. The scene flow $s{f_i}$ for each point ${p_i}$ is a weighted estimate by the information of near points. The non-correspondence complies with the characteristic of two consecutive point clouds captured by LiDAR. The experiments on the real LiDAR scans demonstrated the performance of our network in non-correspondence point clouds.

\section{Hierarchical Attention Learning of Scene Flow in 3D Point Clouds}

\begin{figure*}[t]
	\centering
	\includegraphics[scale=0.65]{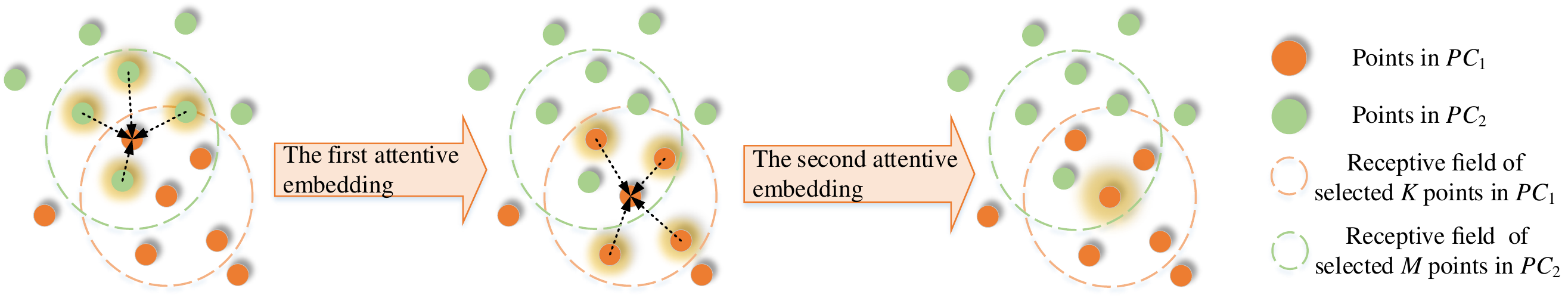}
	\vspace{-0mm}
	\caption{\textbf{The embedding relationship diagram in our double attentive flow embedding in a patch-to-patch manner.
		} 
		The details about the two attentions are presented in Fig. \ref{fig:attention}. The arrows indicate the directions and positions for each embedding.}
	\label{fig:embedding}
\end{figure*}

\begin{figure*}[t]
	\centering
	\includegraphics[scale=0.58]{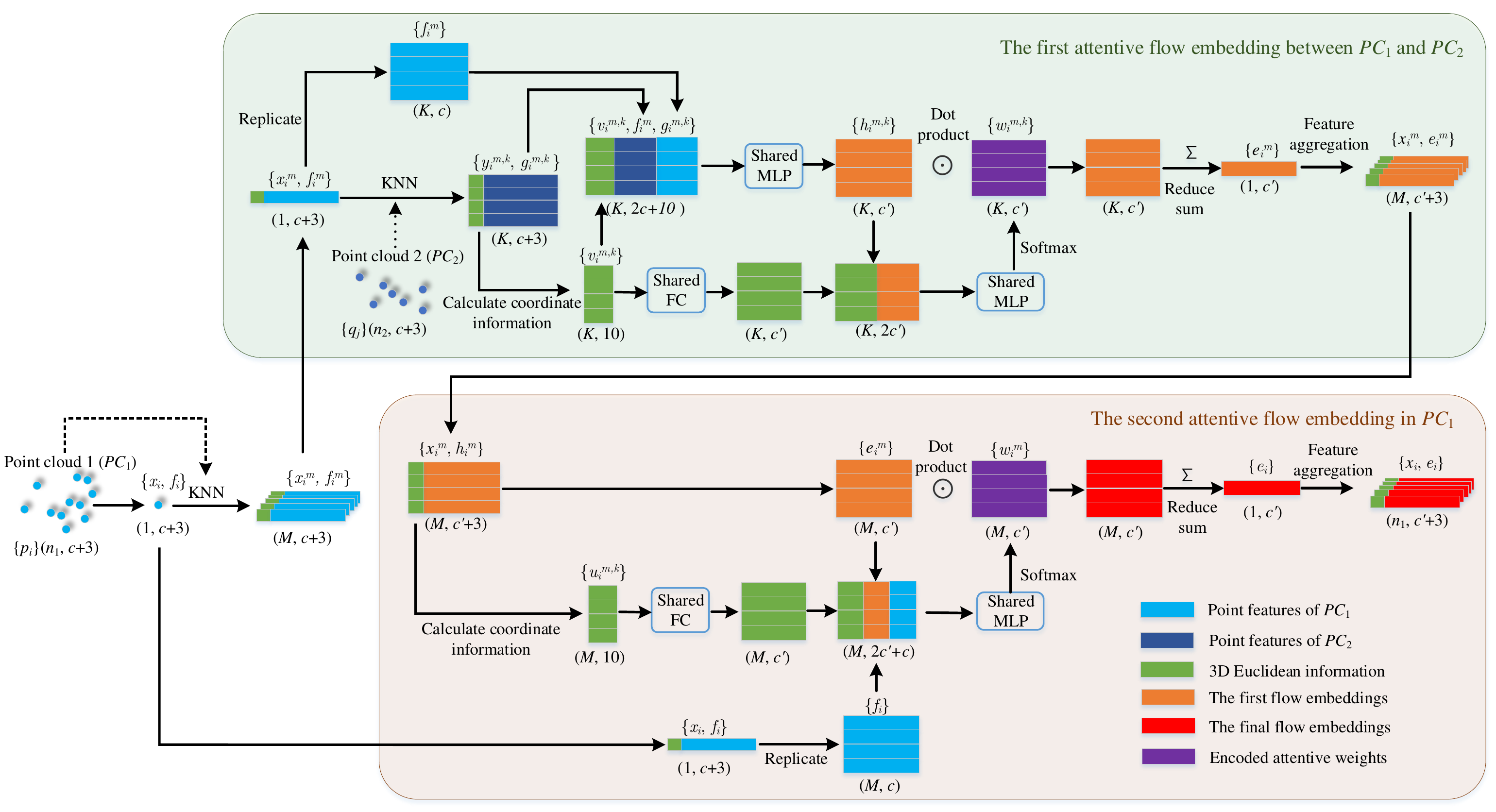}
	\vspace{-4mm}
	\caption{\textbf{The detailed calculation diagram for our double attentive flow embedding method. 	
		} 
	The detailed process description is in Section IV-B.}
	\label{fig:attention}
\end{figure*}

In this section, the proposed hierarchical attention learning network of scene flow is introduced. The network architecture is as Fig. \ref{fig:network}. And there are three basic layers: set conv layer, double attentive embedding layer, and set upconv layer as in Figs. \ref{fig:setconv} and \ref{fig:embedding}. The details about the novel double attentive embedding layer for point mixture are as in Fig. \ref{fig:attention}. The three layers form three main modules: hierarchical point feature encoding, attentive point mixture, and hierarchical attentive flow refinement. Then, we will introduce each module detailly in the following subsections. Finally, the overall architecture of the proposed network is clarified.

\subsection{Hierarchical Point Feature Encoding}

The point clouds from the two frames are firstly encoded individually. For each point cloud, encoding process is hierarchical, which also brings benefits to the hierarchical refinements of scene flow by skip connections. More details about the skip connections from the point feature encoding module to the flow refinement module are in Sections IV-C.1 and IV-C.4. 

For the hierarchical point feature encoding of point clouds, we adopt the set conv layer in PointNet++ \cite{qi2017pointnet++}. Because the point clouds are irregularly distributed in 3D space, the point features are gathered and encoded by considering Euclidean distance in each layer as in Fig. 1 (a). For a set conv layer, the input is $n$ points $\{{p_i} = \{{x_i},{f_i}\} |i = 1, \ldots ,n\} $, where each point ${p_i}$ has the 3D coordinate ${x_i}\in{\mathbb{R}^3}$ and the feature  ${f_i}\in{\mathbb{R}^c}$ ($i = 1, \ldots ,n$). The output of the layer is $n'$ $(n' < n)$ points $\{{p'_j} = \{{x'_j},{f'_j}\} |i = 1, \ldots ,n'\}$, where each point ${p'_j}$ has the 3D coordinate ${x'_j} \in {\mathbb{R}^3}$ and the feature ${f'_j}\in{\mathbb{R}^{c'}}$ ($j = 1, \ldots ,n'$). Specially, the $n'$ points are sampled from the input $n$ point through random sampling in the first set conv layer and Farthest Point Sampling (FPS) in the later set conv layers. (The previous work \cite{liu2019flownet3d} uses FPS in the first layer. The reason for using random sampling in our first layer will be explained in Section IV-D.) The 3D coordinate ${x'_j}$ of each sampled point ${p'_j}$ is output directly, but the feature ${f'_j}$ is extracted as follows. 

For each ${p'_j}$, $K$ Nearest Neighbors (KNN)  $\{ p_j^k = \{x_i^k,f_i^k\}|k = 1, \ldots ,K\} $ are selected from the $n$  input points, where each neighbor  $p_j^k$ has 3D coordinate $x_i^k$ and feature $f_i^k$.  Next, the shared Multi-Layer Perceptron (MLP) with learnable parameters and max-pooling are applied to learn the local feature  ${f'_j} \in {\mathbb{R}^{c'}}$ from the selected $K$ points for each sampled ${p'_j}$. Specially, the shared MLP applied to the $K$ points realizes the feature dimension from $c + 3$ (concatenated $x_j^k - {x'_j}$ and ${f_j}$) to $c'$. Then the max-pooling applied to each feature dimension of the  $K$ points gathers the $K$ point features to one local aggregated point feature. The formula is:
\begin{equation}
{f'_j} = \mathop {{\text{maxpool}}}\limits_{k = 1,...,K} ({\text{MLP}}(x_j^k - {x'_j}) \oplus f_j^k),
\end{equation}
where  $ \oplus $ indicates the concatenation of two vectors. The maxpool means the max-pooling operation.

\subsection{Attentive Point Mixture }

In order to estimate the scene flow in non-corresponding point sets, a weighted flow embedding from two sets of points is needed, which is to estimate the target position of each flowing point in $PC_1$ through some sampled neighbor points in $PC_2$. In order to obtain larger receptive field for each point in the first frame and utilize the structural information in 3D Euclidean space of two sets of points simultaneously, we adopt a patch-to-patch manner \cite{wu2019pointpwc}, shown in Fig. \ref{fig:embedding}, to learn motion encoding from two sets of points. In addition, we expect to pay more attention on matched regions and features to find the right flowing direction. Therefore, a novel double attentive embedding layer is proposed for point mixture. 

The input of the attentive embedding layer is two sets of point clouds: $P{C_1} = \{ {p_i} = \{ {x_i},{f_i}\} |i = 1, \ldots ,{n_1}\} $ and $P{C_2} = \{q_j = \{y_j,g_j\}|j = 1, \ldots ,{n_2}\}$. The  ${x_i},{y_j} \in {\mathbb{R}^3}$ represent the 3D coordinates and ${f_i},{g_j} \in {\mathbb{R}^c}$ represent the features for points ${p_i},{q_j}$. The layer outputs $\mathcal{O} = \{o_i = \{x_i,e_i\}|i = 1, \ldots ,n_1\}$, where ${e_i}\in{\mathbb{R}^{c'}}$ is a learned flow embedding that encodes the point motion for the point ${p_i}$ in $P{C_1}$. 

Fig. \ref{fig:embedding} briefly presents the embedding process, each point ${p_i}$ selects $M$ nearest neighbors ${\mathcal{P}_i} = \{p_i^m = \{ x_i^m,f_i^m\} | m = 1, \ldots ,M\} $  in $P{C_1}$, and each point $p_i^m$ selects $K$ nearest neighbors $\mathcal{Q}_i^m = \{q_i^{m,k} = \{y_i^{m,k},g_i^{m,k}\} | k = 1, \ldots ,K\}$ in $P{C_2}$. The $K$ points $\mathcal{Q}_i^m$ in $P{C_2}$ are first used to encode the point motion. The encoded information is embedded into $p_i^m$ and updates the point feature $f_i^m$ to first flow embedding $e_i^m$ $(m = 1, \ldots ,M)$. Then, the obtained $M$ flow embeddings ${\mathcal{O}_i} = \{ o_i^m = \{ x_i^m,e_i^m\} |m = 1, \ldots ,M\}$ are aggregated into ${p_i}$ $(i = 1, \ldots ,{n_1})$ and update the point feature $f_i$ to final flow embeddings $e_i$. And the final outputs are $\mathcal{O} = \{o_i= \{x_i,e_i\} | i = 1, \ldots ,n_1\}$. The details about the two attentive embeddings are presented in Fig. \ref{fig:attention}. 

For the first attentive embedding: $p_i^m = \{x_i^m,f_i^m\} ,\{q_i^{m,k} = \{ y_i^{m,k},g_i^{m,k}\} \} _{k = 1}^K \to r_i^m = \{x_i^m,e_i^m\}$ $(i = 1, \ldots ,{n_1};m = 1, \ldots ,M)$, the 3D Euclidean space information is first calculated as follows:
\begin{equation}
v_i^{m,k} = x_i^m \oplus {y}_i^{m,k} \oplus (x_i^m - {y}_i^{m,k}) \oplus \left\| {x_i^m - {y}_i^{m,k}} \right\|,
\end{equation}
where $\left\| \cdot \right\|$ indicates the Euclidean distance between two 3D coordinates. The first flow embedding uses the features and 3D Euclidean space information from the two frames of point clouds. And the formula to calculate the first flow embeddings before attentive weighting is:
\begin{equation}
h_i^{m,k} = {\text{MLP}}(v_i^{m,k} \oplus g_i^{m,k} \oplus f_i^m),
\end{equation}
where $h_i^{m,k} \in {\mathbb{R}^{c'}}$. Moreover, the spatial structure information not only helps to determine the similarity of points, but also can contribute to deciding soft aggregation weights of the queried points, so 3D Euclidean information $v_i^{m,k} \in {\mathbb{R}^{10}}$ is also used in the attentive weight learning. The first attentive weight is:

\begin{equation}
w_i^{m,k} = {\text{softmax(MLP(FC(}}v_i^{m,k}) \oplus w_i^{m,k})),
\end{equation}
where a fully connected layer (FC) is used for point position encoding, and softmax activation function is used to normalize the attention. The first attentive flow embedding located at $x_i^m$ is:
\begin{equation}
e_i^m = \sum\limits_{k = 1}^K {h_i^{m,k} \odot w_i^{m,k}},
\end{equation}
where $\odot$ means dot product. That is, the flow embeddings located at the $M$ locations $\{x_i^m|m = 1, \ldots ,M\}$ are ${\mathcal{O}_i} = \{o_i^m = \{ x_i^m,e_i^m\} |m = 1, \ldots ,M\}$.

Next, for the second attentive embedding:
${p_i} = \{ {x_i},{f_i}\}, \{ r_i^m = \{ x_i^m,e_i^m\}\}_{m = 1}^M \to o_i = \{x_i,e_i\}$ $(i = 1, \ldots ,{n_1})$, the 3D Euclidean space information is first calculated as follows:
\begin{equation}
v_i^m = {x_i} \oplus x_i^m \oplus ({x_i} - x_i^m) \oplus \left\| {{x_i} - x_i^m} \right\|.
\end{equation}
For the second embedding, the key is to decide how to weightly aggregate the first flow embeddings by the features of central point and 3D local geometry strcture. So, the second attentive weight for the $e_i^m$ is:

\begin{equation}
w_i^{m} = {\text{softmax(MLP(FC(}}v_i^m) \oplus e_i^m \oplus f_i)),
\end{equation}
The final attentive flow embedding located at ${x_i}$ is:
\begin{equation}
{e_i} = \sum\limits_{m = 1}^M {e_i^m \odot w_i^{m}}.
\end{equation}

The use of multiple 3D Euclidean information is inspired by the learning of 3D semantic segmentation \cite{hu2020randla}. Based on the insight that the scene flow is between each pair of patch points, our double attentive embedding method applies MLP to the preliminary concatenated features before the attention, while in \cite{hu2020randla}, it is the weighted information that enters the MLP. In addition, because the features before the attentive weighting have been encoded as matching information and lose the original spatial information, we reuse the spatial information to achieve soft weighting. The comparative experiment and ablation study verify the effectiveness of our designs in Section V.

Similar with \cite{liu2019flownet3d}, the set conv layer is used after our double attentive flow embedding process for flow encoding. To emphasize its effect, we name it flow feature encoding layer here. The further flow encoding mixes flow features in larger receptive fields, which makes the indistinguishable objects obtain more surrounding information, and also makes the scene flow smoother. The ablation studies in Section V-D.4 demonstrate the gains of further flow encoding.

\subsection{Hierarchical Attentive Flow Refinement}

The hierarchical attentive flow refinement module is designed to propagate and refine scene flow layer by layer, where the key componet is the flow refinement layer. As shown in Fig. \ref{fig:network} (b), each flow refinement layer has four key components: set upconv layer, position updating layer, attentive flow re-embedding layer, and flow correcting layer. They are introduced and organized as follows:

\subsubsection{Set Upconv Layer} 
\begin{figure*}[t]
	\centering
	\includegraphics[scale=0.6]{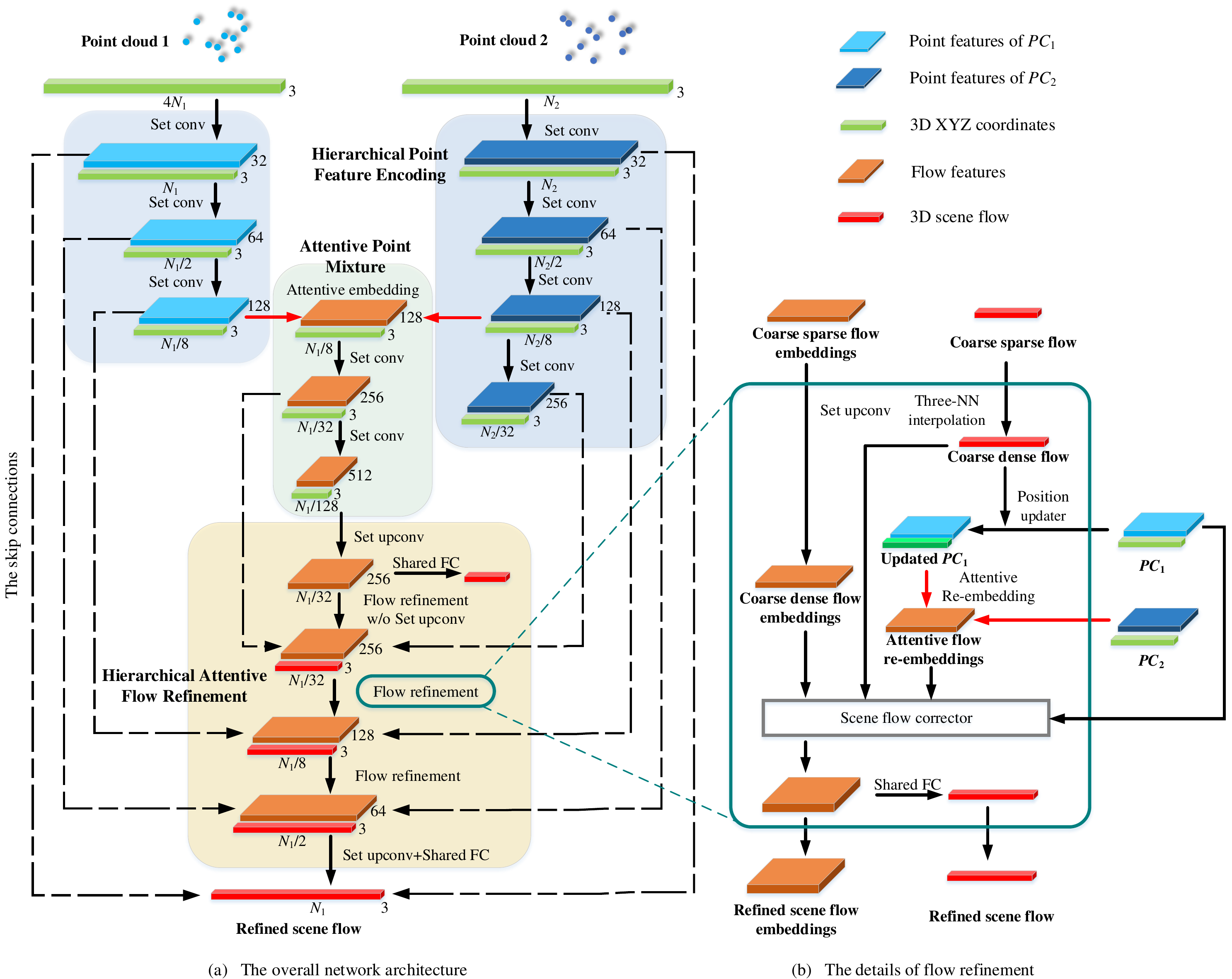}
	\vspace{-4mm}
\caption{\textbf{The architecture of our proposed network for scene flow estimation. 
	} Given two frames of point clouds, the network learns to predict the scene flow from coarse to fine. And the key modules are introduced in Section IV.}
	\label{fig:network}
\end{figure*}

The set upconv layer \cite{liu2019flownet3d} is used to obtain coarse dense flow embeddings from spare flow embeddings as our first step of flow refinement. As shown in Fig. \ref{fig:setconv} (b), the inputs of set upconv layer are sparse $n$ points with flow embeddings $\{p_i = \{x_i,f_i\} |i = 1, \ldots ,n\}$ and the dense $n'$ $(n' > n)$ point coordinates $\{ {x'_j}|j = 1,2,...n'\} $. The outputs are dense $n'$  points with flow embeddings $\{p'_j = \{x'_j,f'_j\} |j = 1, \ldots ,n'\}$. This layer propagates flow features from the sparse point locations to the dense point locations through a learnable KNN feature aggregating manner. It is similar with set conv layer but the aggregated point coordinates are not sampled from input points. The aggregated point coordinates are directly from the original dense point coordinates from hierarchical point feature encoding module. The skip connections in Fig. \ref{fig:network} (a) show which layers are connected. Like \cite{liu2019flownet3d}, the skip connection not only indicates the aggregation positions, but also means concatenating the output features of set conv layers. 

The obtained coarse dense flow embeddings are the preliminary embeddings needed to be further refined in each flow refinement process. 

\subsubsection{Position Updating Layer} In order to make flow re-embedding and flow correction latter, the dense point positions in $PC_1$ are first updated through the coarse dense flows $\{s{f_i}|i = 1, \ldots,n\}$. The coarse dense flows are obatined by interpolation from the three nearest neighbors (Three-NN) of the input coarse sparse flow.

The original points are $PC_1=\{p_i = \{ x_i,f_i\} |i = 1, \ldots,n\} $. Then, the coarse dense flows are added to the original coordinates of the points. So, the updated point clouds are $PC'_1=\{p'_i = \{x'_i,f_i\} |i = 1, \ldots ,n\}$, where $x'_i = x_i + sf_i$.

\subsubsection{Attentive Flow Re-embedding Layer}
Based on the updated point clouds in the first frame after position updating layer and the point clouds in the second frame, the flow re-embeddings are generated by the same method with the attentive flow embedding layer as in Section IV-B. We refer to it as ``re-embedding'' to highlight the role of optimizing the initial flow embeddings in the flow refinement module. And the output flow re-embeddings of this layer are denoted as $RE = \{re_i|re_i \in {\mathbb{R}^c}\} _{i = 1}^n$. The flow re-embeddings are about the updated $PC'_1$ and $PC_2$, so the flow re-embeddings denote the ``residual flow'' after the coarse flow. The residual part of the flow and the coarse flow are input to the following flow corrector layer for flow optimizing.

\subsubsection{Flow Correcting Layer}
The flow correcting layer is to correct the flow embeddings. The operation of this layer is a shared MLP with four inputs: coarse dense flow embeddings $E = \{e_i | e_i \in {\mathbb{R}^c}\}_{i = 1}^n$ (output of set upconv layer in Section IV-C.1), up-sampled coarse dense flow $SF = \{sf_i|sf_i \in {\mathbb{R}^3}\}_{i = 1}^n$ (obtained in position updating layer in Section IV-C.2), the flow re-embeddings $RE = \{re_i|re_i \in {\mathbb{R}^c}\} _{i = 1}^n$ (output of the attentive flow re-embedding layer in Section IV-C.3), and the point features of the first frame  $F = \{f_i|f_i \in {\mathbb{R}^c}\} _{i = 1}^n$ (skip linked from the hierarchical point feature encoding module). The MLP with learnable weights can optimize the coarse dense flow embeddings. And the output is the refined flow embeddings ${E'} = \{e'_i | e'_i \in {\mathbb{R}^{c'}}\}_{i = 1}^n$. The formula is:

\begin{equation}
e'_i = {\text{MLP}}(e_i \oplus sf_i \oplus f_i \oplus re_i).
\end{equation}
Then the refined scene flow $sf'_i$ is obtained by a shared FC on the refined flow embeddings $e'_i$ as follows:
\begin{equation}
	sf'_i = {\text{FC}}(e'_i).
\end{equation}

\begin{figure*}[t]
	\centering
	\includegraphics[scale=0.32]{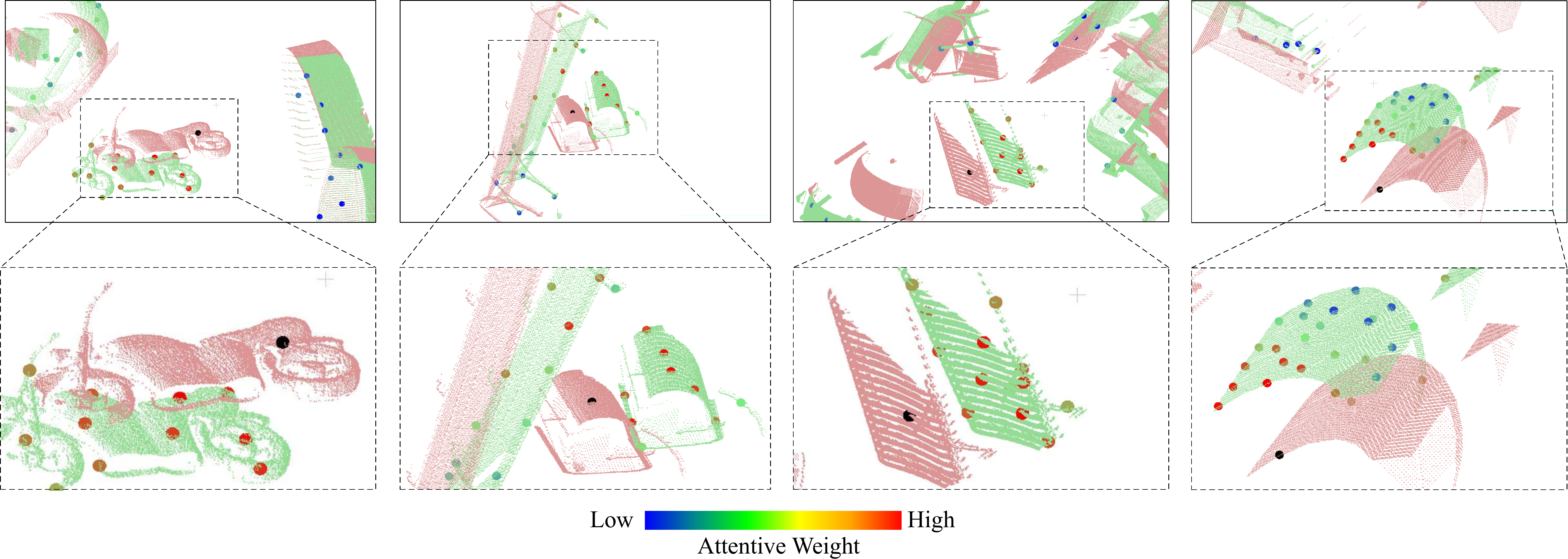}
	\vspace{-1mm}
	\caption{\textbf{Visualization of the attention map. 
		} We show three examples of attention maps on the queried points generated by our trained model. The red small points are $P{C_1}$, and the green small points are $P{C_2}$. The big black point in each $P{C_1}$ is the embedded central point. The big points with other color in $P{C_2}$ are the queried points, whose color denotes the magnitude of the attentive weights. The several queried points actually corresponding to the embedded central point have the heightest attentive weights, which is in line with human intuition to softly estimate the scene flow from the corresponding several points.}
	\label{fig:Attention_Visualization}
\end{figure*}

\subsubsection{The Architecture of a Flow Refinement Layer}
The overview of a flow refinement layer is shown in Fig. \ref{fig:network} (b). The input is coarse flow embeddings, the coarse sparse flow and the point clouds of two frames (skip linked from the hierarchical point feature encoding module); the output is the refined scene flow embeddings and refined scene flow. 

First, the coarse sparse flow embeddings are input to the set upconv layer and coarse dense flow embeddings are obtained. Coarse dense flow is interpolated from the input coarse sparse flow. Then, the $P{C_1}$ is updated through the position updater. The updated $P{C'_1}$ and $P{C_2}$ are used for attentive flow re-embedding and the flow re-embeddings are obtained. Next, the coarse dense flow embeddings, the coarse dense flow, the flow re-embeddings, and the features of $P{C_1}$ are input to scene flow corrector and the refined scene flow embeddings are output. At last, refined scene flow is obtained by a FC on the refined scene flow embeddings.

\subsubsection{The Architecture of the Hierarchical Attentive Flow Refinement module}
For the first flow refinement, the initial coarse dense flow is generated by a shared FC applied to the first coarse dense flow embeddings, which is set upconved from the output of attentive point mixture module. Then, there are three flow refinement layers. At last, in order to save GPU memory, a set upconv layer and a shared FC layer are used to estimate the final refined scene flow.

\subsection{The More-for-Less Hierarchical Attention Network Architecture}
The whole network architecture is shown in Fig. \ref{fig:network} (a). The inputs of the network are point clouds $P{C_1}$ and $P{C_2}$ from two consecutive frames, the proposed network architecture is composed of three modules: hierarchical point feature encoding, attentive point mixture, and hierarchical attentive flow refinement. The hierarchical point feature encoding module has three set conv layers for $P{C_1}$ and four set conv layers for $P{C_2}$. The same pyramid levels of the point feature encoding module share learnable weights. The point mixture module has one attentive embedding layer and two set conv layers. There are three flow refinement layers and one final flow estimation layer (that is set upconv + shared FC) in the hierarchical attentive flow refinement module. The coarse flow is initialized by an shared FC on the set upconved flow embeddings. Then, the coarse flows are refined layer by layer. All MLPs are with BatchNorm and ReLU activation. 

To balance the precision and resource consumption, our network has different input and output number of points,  as shown in Fig.\ref{fig:network} (a). The scene flow of all input points can be estimated by multiple testing if we use random sampling in the first layer of hierarchical point feature encoding module. The comparative experiments about the network structure details are in Section V-D.
 
\subsection{Discussion on the Attention Visualization}

We use the attentive weights to softly estimate the scene flow. Unlike the previous methods using max-pooling to select the max activation \cite{liu2019flownet3d} or just using the relative position encoding \cite{wu2019pointpwc} for the weighting, we use the attention based on the corelation feature and 3D Euclidean feature to weight the neighbor information twice. The Fig. \ref{fig:Attention_Visualization} shows that the closer point is not necessarily the point with higher attention, and the points from similar structure have higher attentive weights. What's more, the queried points actually corresponding to the central black point have the highest attentive weights, which is in line with human intuition.

\setlength{\tabcolsep}{1.8mm}
\begin{table*}[t]
	\begin{center}
		\caption{Quantitative evaluation results on FlyingThings3D \cite{mayer2016large} and KITTI scene flow datasets \cite{menze2018object}. All methods are trained on FlyingThings3D dataset \cite{mayer2016large}. The top part shows evaluation results on FlyingThings3D dataset \cite{mayer2016large} and the bottom part shows evaluation results on KITTI \cite{menze2018object} (ground removed). The results of FlowNet3D \cite{liu2019flownet3d} come from the implement in \cite{gu2019hplflownet}. The implement of FlowNet3D in \cite{gu2019hplflownet} is based on the same data preprocess and setting with HPLFlowNet \cite{gu2019hplflownet} and better than the results in \cite{liu2019flownet3d}. Lower values are better for the error metrics consisting of EPE3D, Outliers3D, and EPE2D. Higher values are better for the accuracy metrics consisting of Acc3D Strict, Acc3D Relax, and Acc2D.}
		\label{table:flyingthing3d}
		\begin{tabular}{clcccccccc}
			\toprule
			Evaluation Dataset       & Method      &  Training Data  &Input      & EPE3D & Acc3D Strict         & Acc3D Relax & Outliers3D & EPE2D & Acc2D\\ \midrule
			& FlowNet3 \cite{ilg2018occlusions}   &  Quarter  & RGB stereo   & 0.4570                        & 0.4179               & 0.6168                          & 0.6050   & 5.1348 & \bf 0.8125                   \\ \cline{2-10}\noalign{\smallskip}

			& ICP \cite{besl1992method}   &  No    & Points & 0.4062                        & 0.1614               & 0.3038                          & 0.8796   & 23.2280 & 0.2913                    \\
			& FlowNet3D \cite{liu2019flownet3d} &  Quarter  & Points  & 0.1136                        & 0.4125               & 0.7706                          & 0.6016      & 5.9740 & 0.5692                 \\
			& SPLATFlowNet\cite{su2018splatnet}&  Quarter & Points   & 0.1205                        & 0.4197               & 0.7180                          & 0.6187     &  6.9759 & 0.5512                \\
			& HPLFlowNet \cite{gu2019hplflownet}   &   Quarter  & Points      & 0.0804                        & 0.6144               & 0.8555                          & 0.4287        & 4.6723 & 0.6764               \\
			& HPLFlowNet \cite{gu2019hplflownet}   &   Complete  & Points      &       0.0696                 &    ---          &      ---                    & ---     & --- &   ---          \\
			& PointPWC-Net \cite{wu2019pointpwc}   &   Complete  & Points      &       0.0588                &    0.7379          &             0.9276             & 0.3424     & 3.2390 &    0.7994         \\
			\multirow{-9}{*}{\begin{tabular}[c]{@{}c@{}}FlyingThings 3D \\ dataset \cite{mayer2016large} \end{tabular}} &  Ours      &  Quarter & Points   & 0.0511                              &                  0.7808    &     0.9437                            &    0.3093   &    2.8739&         0.8056             \\ 
			&  Ours      &  Complete & Points   &  \bf0.0492                             &   \bf0.7850                   &  \bf0.9468                               &  \bf0.3083      &\bf2.7555 &            0.8111          \\\midrule
			& FlowNet3 \cite{ilg2018occlusions}    &   Quarter & RGB stereo    & 0.9111                        & 0.2039              & 0.3587                         & 0.7463     &  5.1023  & 0.7803                \\
			\cline{2-10}\noalign{\smallskip}
			
			& ICP \cite{besl1992method}   &  No    & Points & 0.5181                        & 0.0669               & 0.1667                          & 0.8712     &  27.6752  & 0.1056                 \\
			& FlowNet3D \cite{liu2019flownet3d}  &   Quarter & Points   & 0.1767                        & 0.3738               & 0.6677                          & 0.5271       &   7.2141  & 0.5093             \\
			& SPLATFlowNet \cite{su2018splatnet} &  Quarter & Points  & 0.1988                        & 0.2174               & 0.5391                          & 0.6575  &   8.2306  &  0.4189                  \\
			& HPLFlowNet \cite{gu2019hplflownet}  &   Quarter    & Points    & 0.1169                        & 0.4783               & 0.7776                          & 0.4103      &   4.8055  &  0.5938                \\
			& HPLFlowNet \cite{gu2019hplflownet}  &  Complete   & Points     &        0.1113                &    ---             &       ---                    &  ---    &   ---   &     ---             \\
			& PointPWC-Net \cite{wu2019pointpwc}  &  Complete  & Points      &        0.0694               &  0.7281              &        0.8884                 & 0.2648   &    3.0062 &       0.7673        \\
			\multirow{-9}{*}{\begin{tabular}[c]{@{}c@{}}KITTI \\ dataset \cite{menze2018object} \end{tabular}}  
			&  Ours      &  Quarter & Points   &  0.0692                             &   0.7532                  &   0.8943                              &   0.2529     &  2.8660 &    0.7811                  \\                                                                                                                                          &  Ours    &  Complete   & Points  & \bf 0.0622         & \bf 0.7649 &     \bf    0.9026                        &       \bf0.2492 &    \bf 2.5140 &   \bf 0.8128               \\ \bottomrule
		\end{tabular}
	\end{center}
\end{table*}

\section{Experiments}

We use a supervised method for the learning of scene flow in 3D point clouds. In this section, the training loss and training details are first introduced. Then, the large-scale synthetic FlyingThings3D dataset \cite{mayer2016large} are used for the training and evaluation as previous studies \cite{liu2019flownet3d,gu2019hplflownet,wu2019pointpwc}, since the ground truth of the scene flow is hard to acquire. Then, the trained model is tested on real LiDAR scans from KITTI scene flow dataset \cite{menze2018object} to demonstrate the generalization ability. In the two datasets, our method is compared with the state-of-the-art methods. At last, several ablation studies are conducted to analyze the key components of the proposed network.

\subsection{Training Details and Evaluation Metrics of the Proposed Network}

\subsubsection{Training Loss}The proposed network is trained in a multi-scale supervised fashion as previous work on optical flow estimation \cite{sun2018pwc,sun2019models} and PointPWC-Net \cite{wu2019pointpwc}. The ground truth scene flow of scale $l$ is denoted as $SF^l = \{ sf_i^l \in {\mathbb{R}^3}|i = 1, \ldots ,N_1^l\}$, and the predicted scene flow of each scale $l$ is denoted as $\widehat {SF^l} = \{ \widehat {sf_i^l} \in {\mathbb{R}^3}|i = 1, \ldots ,N_1^l\}$. Four scales of scene flow are supervised in our network. The training loss is:
\begin{equation}
{L_{loss}} = \sum\limits_{l = 1}^4 {{\alpha _l}\frac{1}{{N_1^l}}\sum\limits_{i = 1}^{N_1^l} {{{\left\| {\widehat {sf_i^l} - sf_i^l} \right\|}_2}}}
\end{equation}

where ${\left\|  \cdot  \right\|_2}$ is the ${L_2}$ norm and  ${\alpha _l}$ is the weight for the $l$ scale loss function. We denote the last refined scene flow as ${l_1}$ scale and ${\alpha _1} = 0.2$, ${\alpha _2} = 0.4$, ${\alpha _3} = 0.8$, ${\alpha _4} = 1.6$.
As shown in Fig. \ref{fig:network}, our network inputs $4{N_1} = 8192$ points and outputs 4 scales of scene flow estimation: $N_1^1 = {N_1} = 2048$ points, $N_1^2 = \frac{1}{2}{N_1} = 1024$ points, $N_1^3 = \frac{1}{8}{N_1} = 256$ points, $N_1^4 = \frac{1}{{32}}{N_1} = 64$ points.

\subsubsection{Implementation Details}In our training process, the input points are randomly sampled from the point clouds of the two frames separately.
The point in first frame may not have a corresponding point in the second frame if random sampling is adopted in each frame separately. The sampling fashion is also used when evaluating in the two datasets. Following the previous works \cite{liu2019flownet3d,gu2019hplflownet,wu2019pointpwc}, the network takes point clouds only containing coordinate information as inputs.
All experiments are performed on a Titan RTX GPU with TensorFlow 1.9.0. The Adam optimizer is used for training with ${\beta _1} = 0.9$, ${\beta _2} = 0.999$. The learning rate is exponential decay with initial learning rate 0.001. The decay step is 40000 and the decay rate is 0.7. The training batch size is 8. 

\begin{figure*}[t]
	\centering
	\includegraphics[scale=0.65]{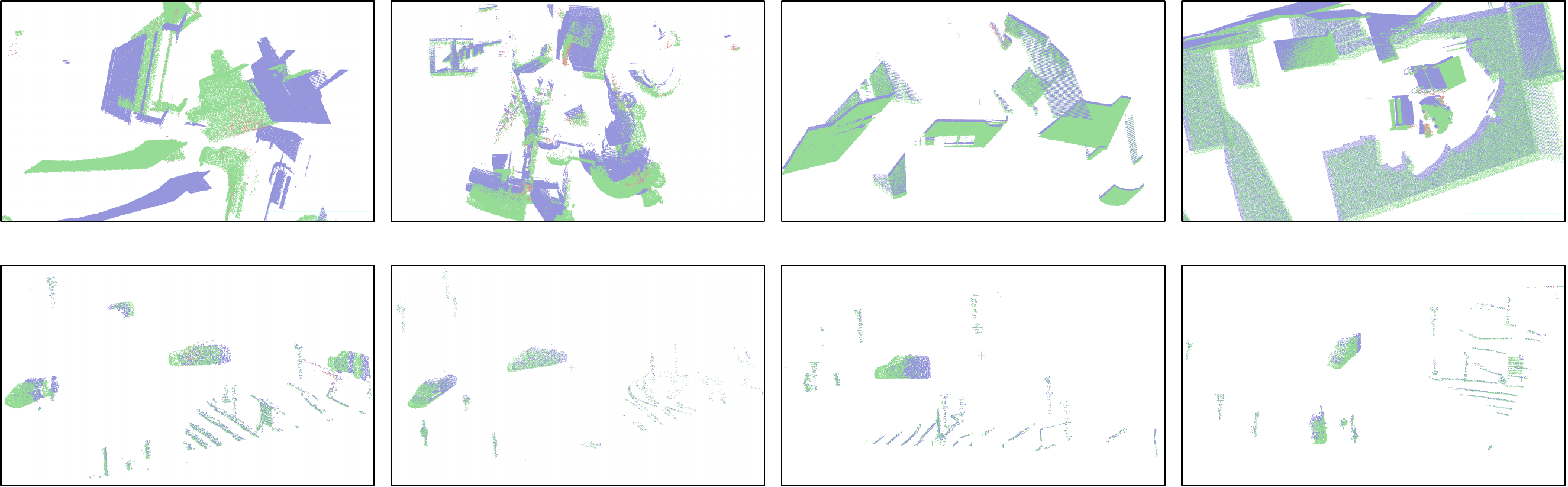}
	\vspace{-1mm}
	\caption{\textbf{Visualization of the accuracy of scene flow estimation.} The top is the results on FlyingThings3D, and the bottom is on KITTI. Blue points are $P{C_1}$. Green points and red points are the predicted flowed points $\widehat {PC_2}$, which is $PC_1+\widehat{SF}$. And the green points are predicted relatively correctly while red points are predicted relatively incorrectly (measured by Acc3D Relax).}
	\label{fig:error}
\end{figure*}

\begin{figure*}[t]
	\centering
	\includegraphics[scale=0.7]{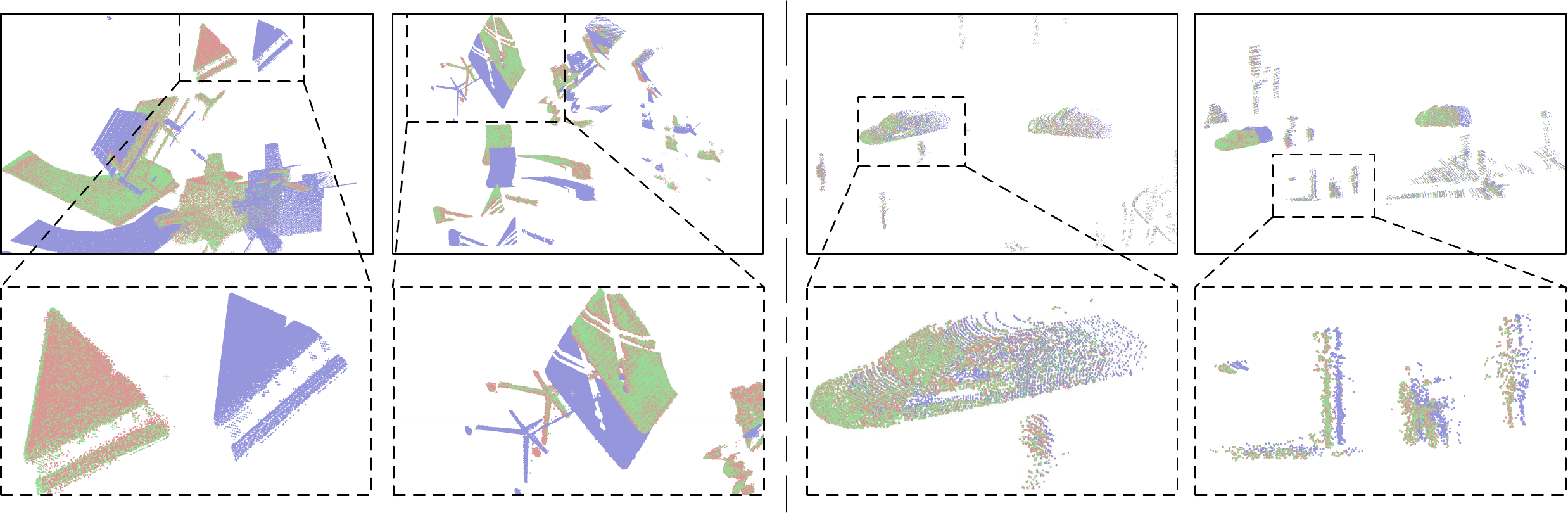}
	\vspace{-3mm}
	\caption{\textbf{Detail visualization of the scene flow estimation.} The left is the results on FlyingThings3D, and the right is on KITTI. Blue points are $PC_1$. Green points are  $PC_2$ and red points are the predicted flowed points $\widehat {PC_2}$.}
	\label{fig:detail}
\end{figure*}

\subsubsection{Evaluation Metrics}For fair comparison, we adopt the evaluation metrics the same as previous works \cite{liu2019flownet3d,gu2019hplflownet,wu2019pointpwc} to evaluate our model. For all evaluations on scene flow estimation, we only evaluate the last refined scene flow output by our network. $SF = \{s{f_i} \in {\mathbb{R}^3}|i = 1, \ldots ,{N_1}\}$ denotes the ground truth scene flow and $\widehat {SF} = \{ \widehat {s{f_i}} \in {\mathbb{R}^3}|i = 1, \ldots ,{N_1}\}$ denotes the predicted scene flow. The evaluation metrics are listed as follows:

The 3D End Point Error \textbf { (EPE3D)} (m): $\frac{1}{{{N_1}}}\sum\limits_{i = 1}^{{N_1}} {\left\| {\widehat {s{f_i}} - s{f_i}} \right\|}$, which measures averaged 3D scene flow estimation error. 

A strict form of scene flow estimation accuracy \textbf{(Acc3D Strict)} : the percentage of $\widehat {s{f_i}}$  such that $\left\| {\widehat {s{f_i}} - s{f_i}} \right\| < 0.05m$  or $\frac{{\left\| {\widehat {s{f_i}} - s{f_i}} \right\|}}{{\left\| {s{f_i}} \right\|}} <5\%$, which represents the percentage of 3D points whose EPE3D $<0.05m$ or relative error $<5\%$.

 A relax form of scene flow estimation accuracy \textbf{(Acc3D Relax)}: the percentage of $\widehat {s{f_i}}$ such that $\left\| {\widehat {s{f_i}} - s{f_i}} \right\|<0.1m$ or $\frac{{\left\| {\widehat {s{f_i}} - s{f_i}} \right\|}}{{\left\| {s{f_i}} \right\|}} <10\%$, which represents the percentage of 3D points whose EPE3D $<0.1m$ or relative error $<10\% $.

 The percentage of 3D outliers \textbf{(Outliers3D)}: the percentage of $\widehat {s{f_i}}$ such that $\left\| {\widehat {s{f_i}} - s{f_i}} \right\| > 0.3m$ or $\frac{{\left\| {\widehat {s{f_i}} - s{f_i}} \right\|}}{{\left\| {s{f_i}} \right\|}} > 10\%$, which represents the percentage of 3D points whose EPE3D $>0.3m$ or relative error $>10\%$.

The 2D End Point Error  \textbf{(EPE2D)} (px): $\frac{1}{{{N_1}}}\sum\limits_{i = 1}^{{N_1}} {\left\| {\widehat {o{f_i}} - o{f_i}} \right\|}$, where the predicted 2D optical flow $\widehat {o{f_i}}$ and the ground truth 2D optical flow $o{f_i}$ are obtained by projecting the initial 3D point clouds and the obtained 3D point clouds by scene flow into 2D image. EPE2D measures averaged 2D optical flow estimation error in pixels.

Optical flow estimation accuracy \textbf{(Acc2D)}: the percentage of $\widehat {o{f_i}}$ such that $\left\| {\widehat {o{f_i}} - o{f_i}} \right\| < 3px$ or $\frac{{\left\| {\widehat {o{f_i}} - o{f_i}} \right\|}}{{\left\| {o{f_i}} \right\|}} < 5\%$, which represents the percentage of 2D points whose EPE2D $<3px$ or relative error $<5\%$.

\subsection{Training and Evaluation on FlyingThings3D}
As the ground truth scene flow is hard to acquire, we adopt the commonly used FlyingThings3D dataset \cite{mayer2016large} (a large-scale synthetic dataset) for training. We also used it for test in this section.
\subsubsection{Experiment Details}Original FlyingThings3D dataset \cite{mayer2016large} is composed of RGB images, disparity map, ground truth optical flow, and occlusion map. There are multiple randomly moving objects in the scene sampled from ShapeNet \cite{chang2015shapenet}. Following the same preprocessing procedure of previous works\cite{gu2019hplflownet,wu2019pointpwc}, 3D point clouds and ground truth scene flow are reconstructed with the 2D data and depth information. Moreover, we also removed the points whose depth is more than $35m$, and only 3D coordinates of points are used. The preprocessed training dataset includes 19,640 pairs of point clouds and the evaluation dataset includes 3,824 pairs. The network is firstly trained on one quarter of the training dataset (4910 pairs) and then the complete training dataset is loaded for finetuning the trained model. The network is evaluated on the whole evaluation dataset for every test (3824 pairs). Other training details are as in Section V-A.

\setlength{\tabcolsep}{2mm}
\begin{table*}[t]
	\begin{center}
		\caption{Ablation studies on the more-for-less network architecture. The bach size is the maximum that a GPU can withstand}
		\label{table:more-for-less}
		\begin{tabular}{c|cccccc|cc}
			\toprule
			Dataset   & Method   & Batch Size            & EPE3D & Acc3D Strict         & Acc3D Relax & Outliers & EPE2D & Acc2D \\ \midrule
			& Replace with Less-for-less    &  8     &   0.0729 
			&    0.5902         &      0.8720                  & 0.4590 &   4.0133 &     0.6698              \\
			&Replace with More-for-more          & 6 
			&    0.0677
			&        0.6161                 &  0.9164 &  0.4290   &   3.6010 &   0.7158           \\

			\multirow{-3}{*}{\begin{tabular}[c]{@{}c@{}}FlyingThings 3D \\ dataset \cite{mayer2016large} \end{tabular}} &  Ours (More-for-less)      & 8     & \bf0.0511                              &                  \bf0.7808    &     \bf0.9437                            &    \bf0.3093   &    \bf2.8739&         \bf0.8056             \\ \bottomrule
		\end{tabular}
	\end{center}
\end{table*}

\setlength{\tabcolsep}{2mm}
\begin{table*}[t]
	\begin{center}
		\caption{Ablation studies on the double attentive embedding method}
		\label{table:attention}
		\begin{tabular}{c|ccccc|cc}
			\toprule
			Dataset   & Method               & EPE3D & Acc3D Strict         & Acc3D Relax & Outliers & EPE2D & Acc2D \\ \midrule
			& Replace with a naive attention \cite{hu2020randla}   &   0.0844
			&0.5495                      &     0.8624                            &  0.5051      &  4.7401  &      0.6548                 \\
			&Ours w/o attention      &   0.0612
			&   0.6982                   &   0.9172                              &  0.3828      &  3.4473  &       0.7524                \\

			\multirow{-3}{*}{\begin{tabular}[c]{@{}c@{}}FlyingThings 3D \\ dataset \cite{mayer2016large} \end{tabular}} & Ours (Double attentive embedding)    & \bf0.0511                              &                  \bf0.7808    &     \bf0.9437                            &    \bf0.3093   &    \bf2.8739&         \bf0.8056             \\  \bottomrule
		\end{tabular}
	\end{center}
\end{table*}

\subsubsection{Comparison with State-of-the-Art}
Table \ref{table:flyingthing3d} shows the quantitative evaluation results on the FlyingThings3D dataset \cite{mayer2016large} with the evaluation metrics introduced in Section V-A. Like HPLFlowNet \cite{gu2019hplflownet}, our method is also compared with FlowNet3 \cite{ilg2018occlusions}, which use the stereo inputs. Extremely wrong predictions are removed for the results of FlowNet3 \cite{ilg2018occlusions} such as opposite disparity estimation \cite{gu2019hplflownet}. FlowNet3 \cite{ilg2018occlusions} has good performance in 2D metrics but poor in 3D metrics, as it is specially optimized for 2D optical flow and disparity estimation. The optimization in 2D makes 3D metrics sensitive to errors. Compared with FlowNet3D \cite{liu2019flownet3d}, the performance of our method has greatly improved. We believe it is the double attentive embedding method and the multiple embedding in hierarchical refinement that contribute mostly. The benefit of the multiple embedding for refinement has been validated in PointPWC-Net \cite{wu2019pointpwc}. We further verified the contribution of our double attentive embedding method in the ablation studies in Section V-D. SPLATFlowNet is a scene flow estimation implement based on SPLATNet \cite{su2018splatnet} by Gu etc. \cite{gu2019hplflownet}. It does not have refinement layers and ours outperforms it. Trough HPLFlowNet \cite{gu2019hplflownet} used the patch correlation and the multiple fusion of signal at different scales, compared with HPLFlowNet \cite{gu2019hplflownet}, our method still has superiority. The first reason is likely that HPLFlowNet \cite{gu2019hplflownet} interpolated signals from continuous domain onto the discrete domain and lost the 3D Euclidean information. The original input is only 3D position information such that 3D Euclidean information is a key component. The second reason is our beneficial double attentive embedding method. What's more, although both FlowNet3D \cite{liu2019flownet3d} and HPLFlowNet \cite{gu2019hplflownet} have refinement stage, they did not use an explicit multiple flow supervision at different scales. The refinement with multiple flow supervision at different scales is tried in PointPWC-Net \cite{wu2019pointpwc} and the results are improved. As the flow embedding is the key component for representing the correlation of point clouds in the scene flow estimation task, our method adopts novel double attentive flow embedding method and novel more-for-less network design. The results show that ours outperforms PointPWC-Net \cite{wu2019pointpwc} remarkably. Overall, our method outperforms all recent works for 3D point cloud input and outperforms 2D method by a large margin in all 3D metrics.

We not only compared our method with the state-of-the-art methods, but also demonstrated the contributions of each proposed component by the ablation studies in Section V-D.
The qualitative results are shown in Figs. \ref{fig:error} and \ref{fig:detail}. It can be seen that although there are various motions in the scenes, most are estimated accurately.

\setlength{\tabcolsep}{2mm}
\begin{table*}[t]
	\begin{center}
		\caption{Ablation studies on the more 3D coordinate information}
		\label{table:10D}
		\begin{tabular}{c|ccccc|cc}
			\toprule
			Dataset   & Method               & EPE3D & Acc3D Strict         & Acc3D Relax & Outliers & EPE2D & Acc2D \\ \midrule
			& Replace with 3D relative coordinates \cite{liu2019flownet3d,wu2019pointpwc}       &    0.0601
			& 0.6881             &  0.9230                         &  0.3865&   3.2454 &      0.7546              \\

			\multirow{-2}{*}{\begin{tabular}[c]{@{}c@{}}FlyingThings 3D \\ dataset \cite{mayer2016large} \end{tabular}} & Ours (10D calculated coordinate information)   & \bf0.0511                              &                  \bf0.7808    &     \bf0.9437                            &    \bf0.3093   &    \bf2.8739&         \bf0.8056             \\  \bottomrule
		\end{tabular}
	\end{center}
\end{table*}

\setlength{\tabcolsep}{2mm}
\begin{table*}[t]
	\begin{center}
		\caption{Ablation studies on the embedding level in point mixture module}
		\label{table:embedding level}
		\begin{tabular}{c|ccccc|cc}
			\toprule
			Dataset   & Method               & EPE3D & Acc3D Strict         & Acc3D Relax & Outliers & EPE2D & Acc2D \\ \midrule
			&  Replace with least point density embedding   &   0.0626
			& 0.6600             &  0.9134                         &0.4179 &3.3706 &  0.7428                   \\

			\multirow{-2}{*}{\begin{tabular}[c]{@{}c@{}}FlyingThings 3D \\ dataset \cite{mayer2016large} \end{tabular}}  & Ours (Middle point density embedding)    & \bf0.0511                              &                  \bf0.7808    &     \bf0.9437                            &    \bf0.3093   &    \bf2.8739&         \bf0.8056             \\  \bottomrule
		\end{tabular}
	\end{center}
\end{table*}

\setlength{\tabcolsep}{2mm}
\begin{table*}[t]
	\begin{center}
		\caption{Ablation studies on the more levels supervised}
		\label{table:supervised}
		\begin{tabular}{c|ccccc|cc}
			\toprule
			Dataset   & Method               & EPE3D & Acc3D Strict         & Acc3D Relax & Outliers & EPE2D & Acc2D \\ \midrule
			& Replace with the least point level supervised   & 0.1226
			&    0.3403      &    0.7015                  &         0.7012      &       6.3487                    &       0.4731               \\

			\multirow{-2}{*}{\begin{tabular}[c]{@{}c@{}}FlyingThings 3D \\ dataset \cite{mayer2016large} \end{tabular}}  & Ours (The least point level not supervised)    & \bf0.0511                              &                  \bf0.7808    &     \bf0.9437                            &    \bf0.3093   &    \bf2.8739&         \bf0.8056             \\  \bottomrule
		\end{tabular}
	\end{center}
\end{table*}

\subsection{Evaluation on the Real LiDAR Scans in KITTI}

In this subsection, to analyze the proposed model’s generalization ability to unseen real-world data, the model trained on FlyingThings3D \cite{mayer2016large} is directly evaluated in real Lidar scans from KITTI \cite{menze2018object} without any finetuning.

\subsubsection{Experiment Details}KITTI Scene Flow 2015 \cite{menze2018object} is the common dataset for scene flow evaluation of RGB stereo based methods. It is composed of 200 training scenes and 200 test scenes. In order to evaluate the points based methods and execute fair comparison with the state-of-the-art methods, we preprocess the data with the same steps as previous works\cite{liu2019flownet3d,gu2019hplflownet,wu2019pointpwc}. Specially, the 3D point clouds and the scene flow ground truth are recovered from the disparity and optical flow ground truth. The points with depth more than $35m$ are removed like in FlyingThings3D \cite{mayer2016large}. The ground is removed by height less than $0.3m$, since its motion is not useful in autonomous and removing the ground is a common step \cite{dewan2016rigid,ushani2017learning,liu2019flownet3d,gu2019hplflownet,wu2019pointpwc}. The 142 training scenes with available raw 3D data is used for evaluation, since the disparity of test set are not available.

\subsubsection{Comparison with State-of-the-Art}The quantitative evaluation results are shown in Table \ref{table:flyingthing3d}. Overall, our method outperforms all state-of-the-art methods including FlowNet3 \cite{ilg2018occlusions} in 2D metric, though FlowNet3 \cite{ilg2018occlusions} has a good performance in 2D metric on FlyingThings3D dataset \cite{mayer2016large}. The better performance of our method than FlowNet3 \cite{ilg2018occlusions} in 2D metrics on the KITTI \cite{menze2018object} is likely due to the better generalization ability of our 3D method. Our method only needs 3D point coordinate as input and learns the geometric transformation from only 3D Euclidean information, while FlowNet3 \cite{ilg2018occlusions} based on RGB stereo uses color information, which has a bigger gap between the synthetic data and real-world data. 
Some visualization results in Figs. \ref{fig:error} and \ref{fig:detail} show the good generalization effect of our method on KITTI.

\subsection{Ablation Study}

In this subsection, a series of ablation studies are conducted on the components of our method. And these studies analyze the contributions of each component in the network.
For all ablation study, we use \textbf {one quarter of training set (4910 pairs)} for reducing training time and test on the whole evaluation set on FlyingThings3D.

\subsubsection{More-for-Less Network Architecture}The more-for-less architecture means that the input points are 4 times of the predicted points. Specially, we set the input point quantity to be 8192 and the number of predicted points to be 2048. The comparison groups consist of 2048 input for 2048 prediction (less-for-less) and 8192 input for 8192 prediction (more-for-more). The predicted points in the more-for-less architecture can contain more information than that in the less-for-less architecture. At the same time, the more-for-more architecture has more points to refinement, which consumes more GPU memories than our more-for-less architecture. They are all trained on a single Titan RTX GPU with different batch size and the evaluation results are in the Table \ref{table:more-for-less}. Note that the more-for-less has the same batch size with the less-for-less, because the set conv layer occupies few memories while the embedding process occupies many more. Compared with the more-for-more and the less-for-less, our method uses more information and reduces resource consumption, achieving best prediction on one GPU.

\subsubsection{Double Attentive Embedding} We propose the double attentive embedding method in the hierarchical architecture. In order to demonstrate the effectiveness of the proposed attention method. We train and evaluate the network with and without the attention. At the same time, we directly use the LocSE and attentive pooling in semantic segmentation \cite{hu2020randla} to replace our first attentive embedding process as a naive attention-based version for comparison, where we move the MLP of the attentive pooling before the attention in order to get closer to our approach, otherwise the result will be worse. Table \ref{table:attention} shows our attention method improves the performance greatly.

\subsubsection{More 3D Coordinate Information} We use the 10 dimision coordinate information in the embedding as shown in Fig. \ref{fig:attention}, like the semantic segmentation \cite{hu2020randla}, rather than only use the relative coordinates of the queried points \cite{liu2019flownet3d,wu2019pointpwc}. And we found the 10 dimision coordinate information contributes to the better performance in Table \ref{table:10D}. 

\subsubsection{The Suitable Embedding Level in Point Mixture Module} Flownet3D \cite{liu2019flownet3d} embeds two point clouds in a middle point density level. They found that a few more set conv layers after embedding benefits point mixture. However, PointPWC-Net \cite{wu2019pointpwc} discused less about the suitable embedding level in point mixture. It becomes a question about which layer in point feature encoding should be embedded in muti-layer supervision. We studied this. Different from \cite{wu2019pointpwc}, whose initial flow embedding is carried out in the least point density level, ours is in middle point density level. So that the initial flow embeddings can be mixed again for spatial smoothness in the following set conv layers, which achieves better results. The ablation studies in Table \ref{table:embedding level} demonstrated it.

\subsubsection{Whether More Levels Are Supervised, the Better} It is a question whether the scene flow of the least density point level should be estimated and supervised. Our network does not estimate the 3D scene flow from the least density level. We believe the supervised signal on the sparsest point level makes large errors because of the inconspicuous spatial structure. As shown in Table \ref{table:supervised}, it is not ture that the more levels are supervised, the better results can be achieved.

\setlength{\tabcolsep}{1.1mm}
\begin{table*}[t]
	\begin{center}
		\caption{Quantitative experiment results on LiDAR odometry task on KITTI odometry dataset \cite{geiger2013vision}. Here $t_{rel}$ and $r_{rel}$ denote the average translational RMSE (\%) and rotational RMSE ($^{\circ}$/100m) respectively on all measurable subsequrences of 100-800 meters in length. `$^*$' means that the sequence is used for training.}
		\label{table:lidar}
		\begin{tabular}{l|cc|cc|cc|cc|cc|cc|cc|cc|cc|cc|cc}
			\toprule
		    &  \multicolumn{2}{c|}{00$^*$}  &\multicolumn{2}{c|}{01$^*$}      & \multicolumn{2}{c|}{02$^*$} & \multicolumn{2}{c|}{03$^*$} &  \multicolumn{2}{c|}{04$^*$} & \multicolumn{2}{c|}{05$^*$} & \multicolumn{2}{c|}{06$^*$} & \multicolumn{2}{c|}{07} & \multicolumn{2}{c|}{08} & \multicolumn{2}{c|}{09} &\multicolumn{2}{c}{10} \\ 
		    \cline{2-23}\noalign{\smallskip}
		    
		    \multirow{-2}{*}{\begin{tabular}[c]{@{}c@{}}Method \end{tabular}}
             &  $t_{rel}$  & $r_{rel}$   & $t_{rel}$                       & $r_{rel}$               & $t_{rel}$                          & $r_{rel}$   & $t_{rel}$ & $r_{rel}$   & $t_{rel}$                          & $r_{rel}$   & $t_{rel}$ & $r_{rel}$    & $t_{rel}$                          & $r_{rel}$   & $t_{rel}$ & $r_{rel}$ & $t_{rel}$                          & $r_{rel}$   & $t_{rel}$ & $r_{rel}$    & $t_{rel}$ & $r_{rel}$       \\
		     \cline{1-23}\noalign{\smallskip}
			
		   ICP    
		   &6.88&	2.99
		   &11.21&	2.58
		   & 8.21	&3.39
		   &11.07&	5.05
		   & 6.64	&4.02
		   & 3.97&	1.93
		   &\bf1.95	&1.59
		   &5.17	&3.35
		   &10.04	&4.93
		   &6.93	&2.89
		   &8.91	&4.74
		          \\ 
		   
		  
		   Ours      
			&\bf2.89	&\bf1.26
		   &\bf2.37	&\bf0.72
		   &\bf2.64	&\bf1.12
		   &\bf2.79	&\bf2.12
		   &\bf1.74	&\bf0.88
		   &\bf3.01	&\bf1.29
		   &3.23	&\bf1.14
		   &\bf2.92    &\bf1.82
		   &\bf4.13	&\bf1.60
		   &\bf3.05	&\bf1.11
		   &\bf3.62	&\bf1.78
			   \\ \bottomrule
		\end{tabular}
	\end{center}
\end{table*}

\subsection{The Application on LiDAR Odometry}

Most existing methods on LiDAR odometry are based on the variants of ICP \cite{besl1992method}. As there is few corresponded points in adjacent frames for the sparse scans of LiDAR, ICP uses the closest point as the correspondence and iteratively applies Singular Value Decomposition (SVD) multiple times to optimize the 6-DoF pose estimation. It is not only time-consuming but also error-prone. The correct corresponded point can be estimated for each point in $PC_1$ by our scene flow network from adjacent frames. Then, SVD can be directly used once to solve the 6-DoF pose transformation matrix in a close-form and there is no need to assume corespondence. Based on these discussion, we applied our scene flow network to the LiDAR odometry, and compared with the ICP-based method.

\subsubsection{Our Method and Experiment Details}

We use the SVD once to get the 6-DoF pose transformation from the $PC_1$ and $\widehat {PC_2}$, where $\widehat {PC_2} = \{ {x'_i} \in {\mathbb{R}^3}|x'_i = x_i + \widehat{sf_i},i = 1, \ldots ,{N_1}\}$. The scene flow $\widehat {sf_i}$ is estimated by our network from $PC_1$ and $PC_2$. 

The odometry experiments are executed on KITTI odometry dataset \cite{geiger2013vision}, which consists of 22 sequences. The 00-10 sequences have ground truth. We use the sequences 00-06 to train our network and the 07-10 to test. In the training, we augment the data by adding random rotation and translation to the point clouds and the ground truth ego-motion changes accordingly. The ground truth scene flow $sf_i$ is obtained by the formula as follows:
\begin{equation}
{sf_i} = R x_i + t - x_i,
\end{equation}
where $R$ and $t$ are ground truth rotation matrix and translation vector between adjacent frames. And we use the obtained scene flow $sf_i$ to supervise our network.

\subsubsection{Comparison with the ICP-Based Method and Discution} We compare our method with the ICP-based LiDAR odometry. The results in Table \ref{table:lidar} show that we outperform the ICP-based method. 

For our method, the key is estimating the accurate scene flow to get accurate point position $\widehat {PC_2}$ in the second frame for $PC_1$. That is, we change the problem form and get better results than the ICP-based method on LiDAR odometry task. Like the recent visual odometry study based on the perspective of optical flow \cite{min2020voldor}, LiDAR odometry from the perspective of scene flow may be worthy of in-depth study.

\section{Conclusion}
In this paper, a novel hierarchical attention learning method of scene flow is proposed. Inspired by the human intuition, our proposed double attentive flow embedding method focuses on several important points to softly estimate the scene flow. Besides, a novel more-for-less network architecture is proposed to balance the precision and resource consumption. A series of ablation studies demonstrated the effectiveness of our network structure settings. 

Extensive experiments in the common FlyingThings3D dataset \cite{mayer2016large} and KITTI dataset \cite{menze2018object} demonstrated that our method can achieve state-of-the-art performance. The proposed network was also applied to a practical LiDAR odometry task. Our method does not need to assume corresponding points or iterations and outperforms the ICP-based method, which shows the great practical application ability.


%

\ifCLASSOPTIONcaptionsoff
  \newpage
\fi

\bibliographystyle{IEEEtran}  
\bibliography{IEEEabrv,bare_jrnl} 

	\vspace{-10mm}
\begin{IEEEbiography}[{\includegraphics[width=1in,height=1.25in,clip,keepaspectratio]{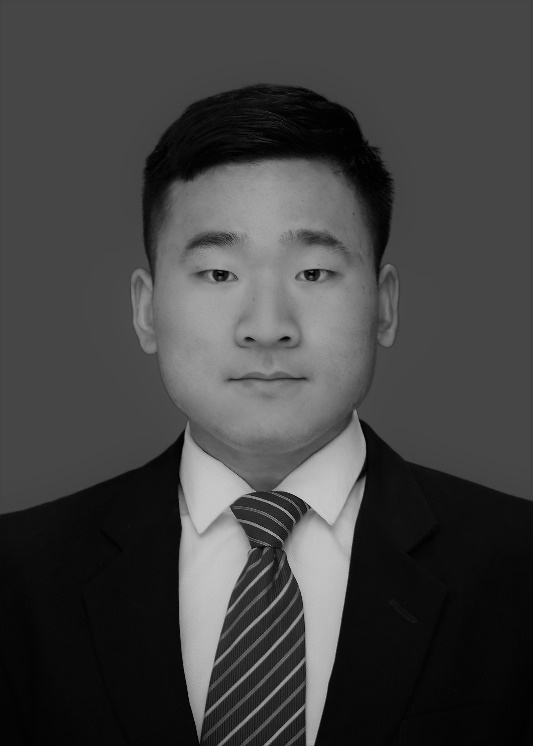}}]{Guangming Wang}
received the B.S. degree from Department of Automation from Central South University, Changsha, China, in 2018. He is currently pursuing the Ph.D. degree in Control Science and Engineering with Shanghai Jiao Tong University. His current research interests include SLAM and computer vision, in particular, 2D optical flow estimation and 3D scene flow estimation.
\end{IEEEbiography}
	\vspace{-10mm}
\begin{IEEEbiography}[{\includegraphics[width=0.9in,height=1.3in,clip]{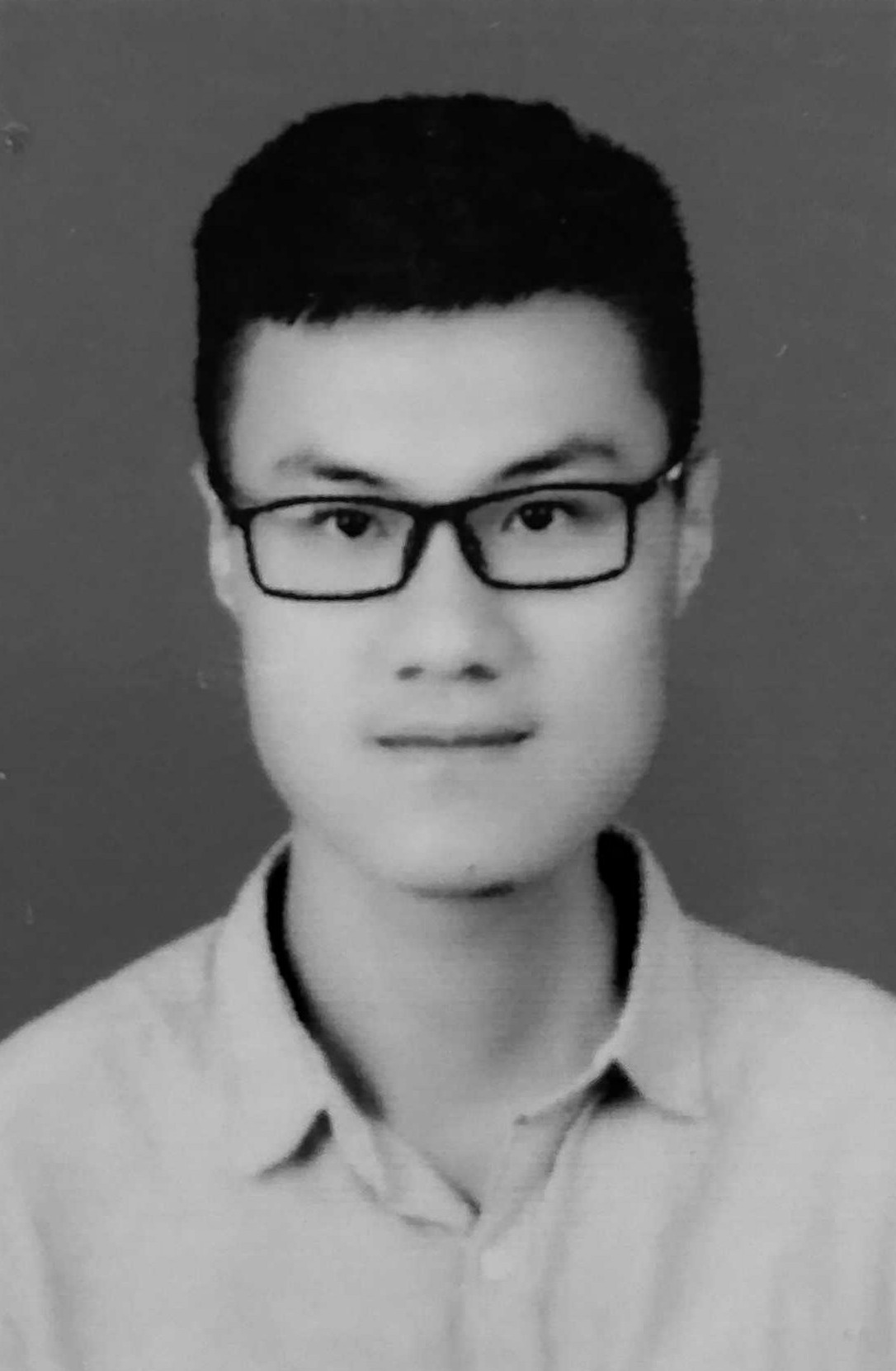}}]{Xinrui Wu}
is currently pursuing the B.S. degree in Department of Automation, Shanghai Jiao Tong University. His latest research interests include SLAM and computer vision. 
\end{IEEEbiography}
	\vspace{-10mm}
\begin{IEEEbiography}[{\includegraphics[width=1in,height=1.25in,clip,keepaspectratio]{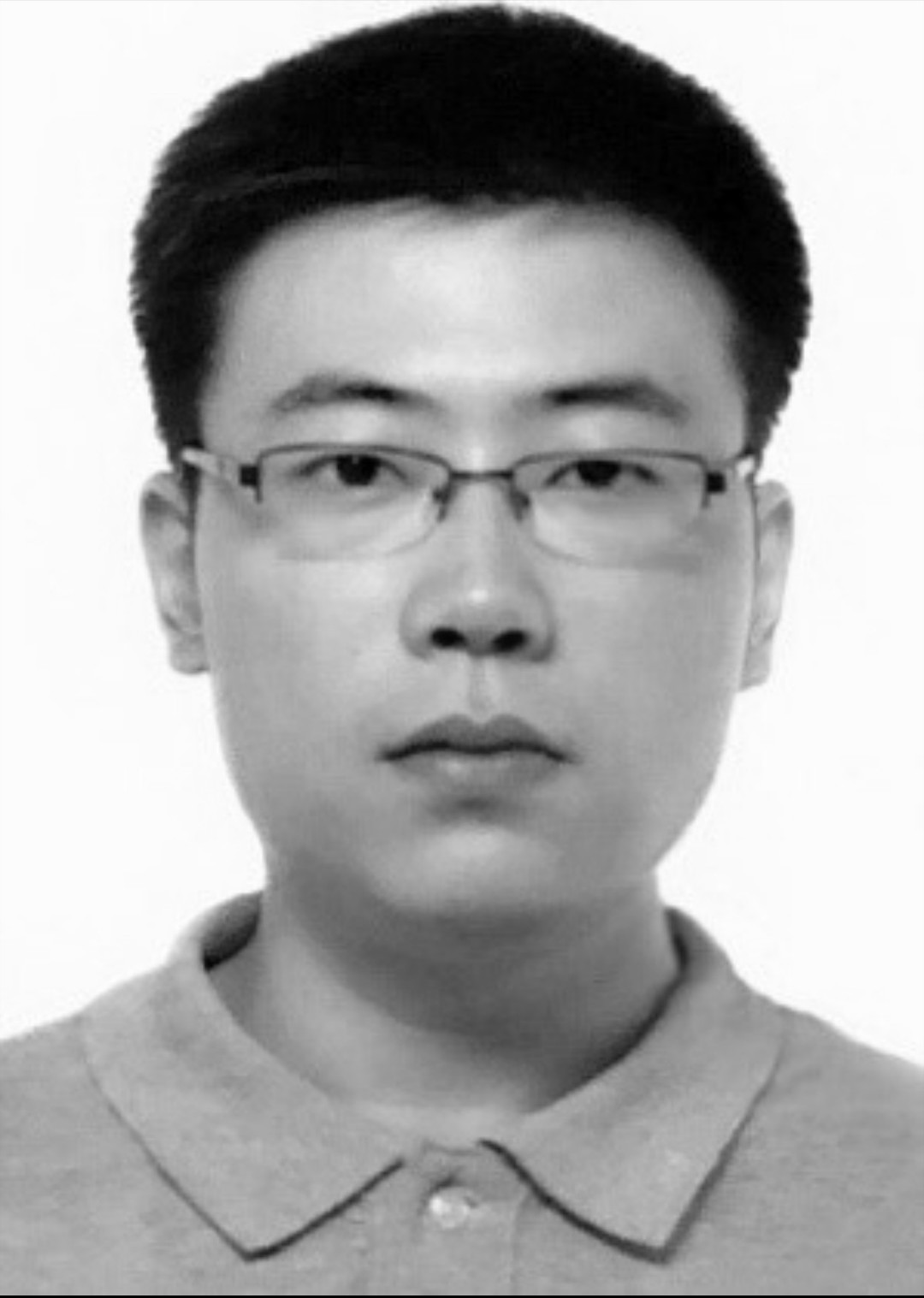}}]{Zhe Liu} received his B.S. degree in Automation from Tianjin University, Tianjin, China, in 2010, and Ph.D. degree in Control Technology and Control Engineering from Shanghai Jiao Tong University, Shanghai, China, in 2016. From 2017 to 2020, he was a Post-Doctoral Fellow with the Department of Mechanical and Automation Engineering, The Chinese University of Hong Kong, Hong Kong. He is currently a Research Associate with the Department of Computer Science and Technology, University of Cambridge. His research interests include autonomous mobile robot, multirobot cooperation and autonomous driving system. 
\end{IEEEbiography}
	\vspace{-10mm}

\begin{IEEEbiography}[{\includegraphics[width=1in,height=1.25in,clip,keepaspectratio]{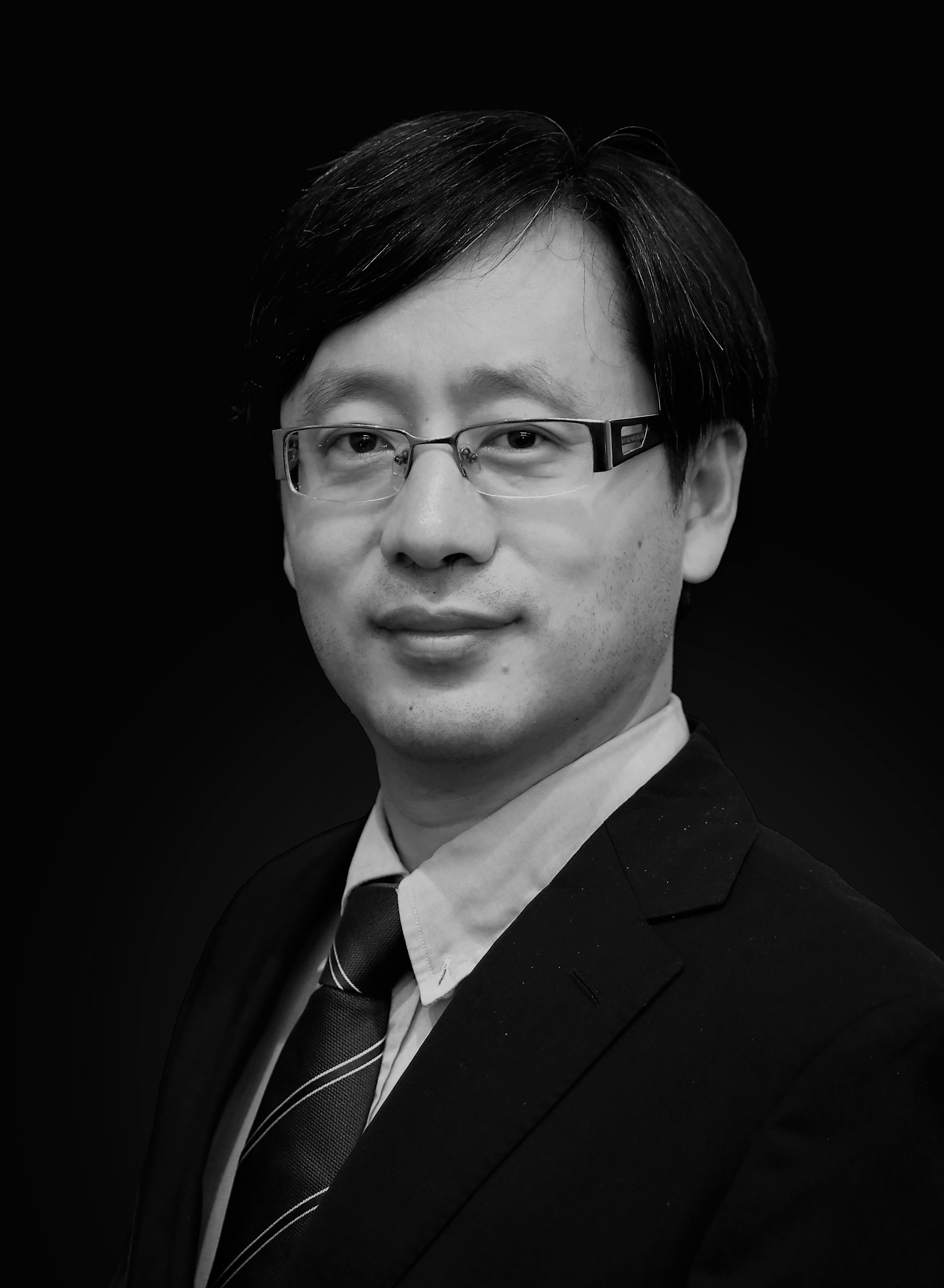}}]{Hesheng Wang}
 received the B.Eng. degree in electrical engineering from the Harbin Institute of Technology, Harbin, China, in 2002, the M.Phil. and Ph.D. degrees in automation and computer-aided engineering from the Chinese University of Hong Kong, Hong Kong, in 2004 and 2007, respectively. From 2007 to 2009, he was a Postdoctoral Fellow and a Research Assistant in the Department of Mechanical and Automation Engineering, the Chinese University of Hong Kong. Since 2009, he has been with Shanghai Jiao Tong University, China, where he is currently a Professor in the Department of Automation. His research interests include visual servoing, service robot, robot control, and computer vision. Dr. Wang is an Associate Editor of Robotics and Biomimetics, Assembly Automation, International Journal of Humanoid Robotics, and IEEE TRANSACTIONS ON ROBOTICS. He is the Program Chair of the 2019 IEEE/ASME International Conference on Advanced Intelligent Mechatronics. 
\end{IEEEbiography}




\end{document}